\documentclass[letterpaper, 10 pt, conference]{ieeeconf}  

\IEEEoverridecommandlockouts                              

\usepackage[english]{babel}

\usepackage{soul} 
\usepackage{marginnote} 

\usepackage{amsfonts}
\usepackage{xcolor}
\usepackage{tcolorbox}
\tcbuselibrary{breakable} 
\newtcolorbox{revisebox}[2][]{
    breakable, 
    colback=blue!5!white, 
    colframe=blue!50!black, 
    title={\textbf{Revisions to Comment~#2}}, 
    fonttitle=\small,
    left=0mm,right=0mm,top=1mm,bottom=1mm, 
    #1 
}

\usepackage{cite}
\usepackage{booktabs}
\usepackage{multirow}
\usepackage[table]{xcolor}
\usepackage{graphicx}
\usepackage{graphicx}
\usepackage{subcaption}
\usepackage{multirow}
\usepackage{booktabs}
\let\labelindent\relax
\usepackage{enumitem}
\usepackage[colorlinks=true, allcolors=blue, hyperfootnotes=true]{hyperref}
\usepackage{utils/symbols}
\usepackage{amsmath}
\usepackage{cuted}
\usepackage{capt-of}

\definecolor{deepblue}{RGB}{31, 78, 121}
\definecolor{darkred}{HTML}{B23A3A}
\newcommand{\method}{\textcolor{darkred}{\textbf{I-BFM}}}
\newcommand{\std}[1]{{\color{gray}{\tiny $\pm #1$}}}
\newcommand{\best}[1]{{\bfseries\boldmath #1}}
\title{\LARGE \bf
\method{}: Reward-Conditioned Robust Humanoid Interaction via Unsupervised Reinforcement Learning}

\author{Ziqi Han$^{1,*}$, Yitang Li$^{2,*}$, Junhan Sun$^{3,*}$, Fanrong Dong$^3$, Yaojie Shen$^4$, Lei Ye$^5$,\\
Zetong Jing$^3$, Yongqi Zhang$^{4,6}$, Yiming Zhang$^4$, Xue Wang$^4$, and Hao Zhao$^{2,4,\dagger}$\\[0.3em]
{\small $^{*}$Ziqi Han, Yitang Li, and Junhan Sun contributed equally.}\\[0.2em]
{\small $^1$Tongji University \quad $^2$Tsinghua University \quad $^3$Zhejiang University}\\
{\small $^4$RoboParty Lab \quad $^5$Harbin Institute of Technology \quad $^6$ShanghaiTech University}\\[0.2em]
{\small $^{\dagger}$Corresponding author.}\\[0.2em]
{\small Project page: \url{https://iamhardworking.github.io/I-BFM/}}}

\begin{document}

\maketitle
\thispagestyle{empty}
\pagestyle{empty}

\setlength{\stripsep}{-0pt}

\begin{strip}
\centering
\includegraphics[width=\textwidth, trim=0cm 0cm 0cm 0cm, clip]{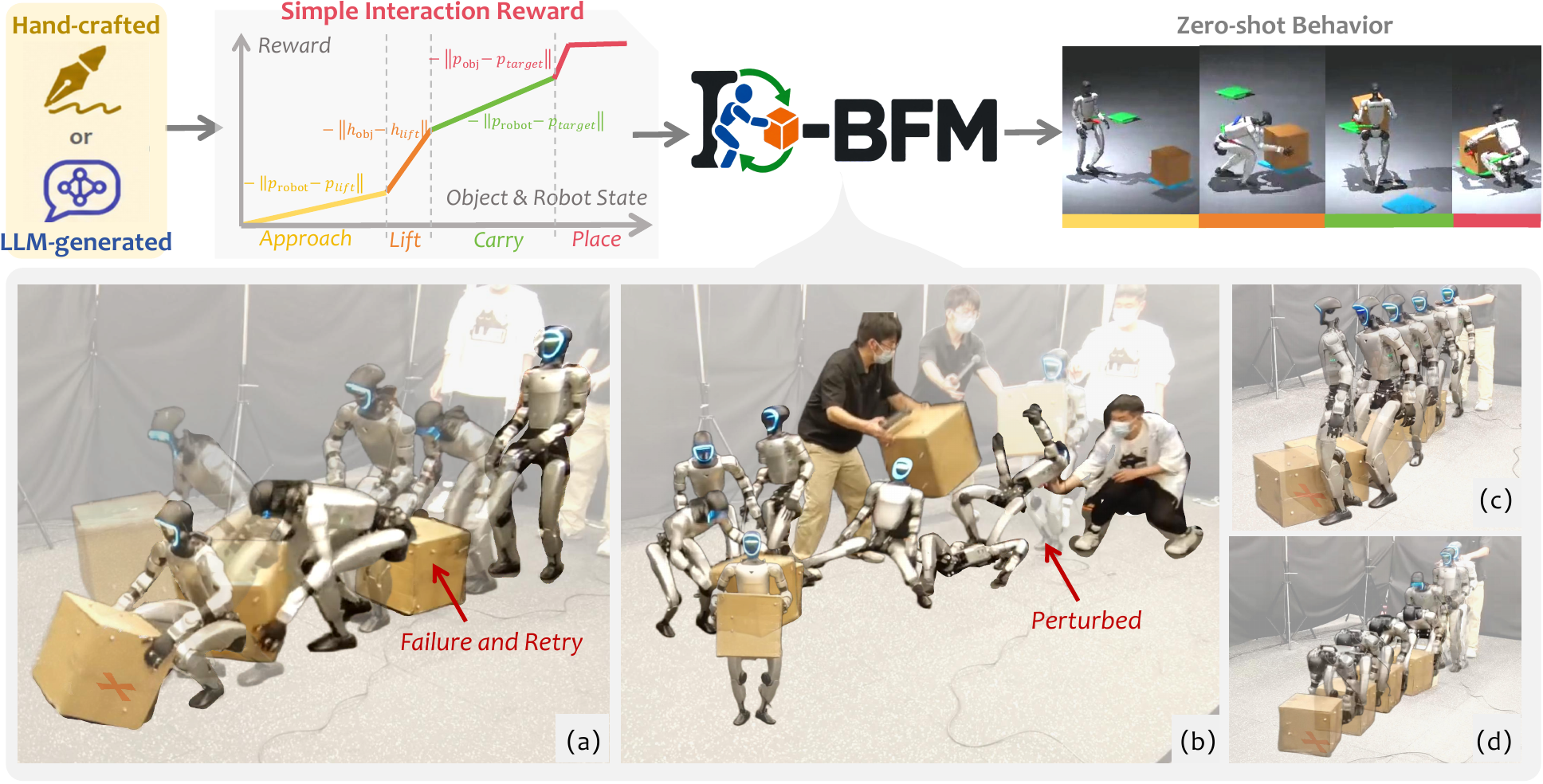}
\captionof{figure}{
\method{}: Reward-Conditioned Robust Humanoid Interaction via Unsupervised Reinforcement Learning.
Given an arbitrary interaction objective through a simple reward interface,
\method{} generates the corresponding behavior zero-shot, without task-specific
retraining.
(a) The robot retries after a failed carrying attempt.
(b) It recovers from an external perturbation and box displacement.
(c) The same policy kicks a box.
(d) The same policy pushes a box.
}
\label{fig:teaser}
\vspace{0.2cm}
\end{strip}

\begin{abstract}
Behavioral foundation models (BFMs) have recently shown that a single humanoid policy can support diverse whole-body control, but extending such generality to physical interaction remains challenging. We introduce \method{}, to our knowledge the first BFM for humanoid–object interaction. Rather than relying on task-specific policies or reference tracking, \method{} learns a shared representation of the coupled dynamics among the humanoid, objects, and their contacts using forward-backward representations and unsupervised reinforcement learning. Given a downstream task reward, the same policy can be directly conditioned on a latent command to execute closed-loop interaction without task-specific policy optimization. To improve interaction control over different time scales, we further train the policy with both short-horizon interaction targets and longer-horizon goal targets. A single \method{} policy performs carrying, pushing, and kicking, while also supporting goal reaching, motion tracking, stylistic control, and long-horizon task chaining. More importantly, it remains effective after large deviations from nominal execution: on Carry, \method{} achieves 94.3\% nominal success and retains 89.3\% success after robot falls, compared with 1.3\% for a planning-based baseline. Real-world experiments on a Unitree G1 further demonstrate diverse loco-manipulation behaviors, rapid recovery from interaction failures and external disturbances, and task chaining without task-specific retraining.
\end{abstract}

\section{INTRODUCTION}

Humanoid robots are envisioned as general-purpose agents capable of performing diverse tasks across different environments. Beyond locomotion, physical interaction with objects is an indispensable capability for realizing this vision. A truly general-purpose humanoid should not only move robustly, but also perform diverse interaction behaviors and effectively recover from unexpected failures during long-horizon tasks.

Recent whole-body foundation models (BFMs) have demonstrated strong generality and robustness in humanoid motion~\cite{bfm0, luo2025sonic, liao2025beyondmimic}. However, extending these capabilities to physical interaction remains challenging. Existing approaches often rely on open-loop teleoperation and behavior cloning, which are difficult to scale and provide limited failure recovery.

Another line of work focuses on humanoid-object co-tracking, where MoCap HOI trajectories are directly used as supervision. While these methods achieve high-quality interaction behaviors, they heavily rely on dense references and are inherently inflexible~\cite{weng2025hdmi,yang2025omniretarget,wang2026humanx}. Some approaches introduce planners~\cite{omnicontact} to generate  intermediate references for the tracker. However, planners introduce additional latency, especially after interaction failures, and constrain behavioral diversity to their predefined rules. More recently, style-learning-based approaches have shown some task-specific generalization, such as transporting boxes between arbitrary configurations, but remain insufficiently general and robust for long-horizon interaction.

These limitations raise a fundamental question: Can we endow a general-purpose humanoid controller with interaction capabilities while preserving its generality and robustness? Our key observation is that these desired properties—including generality, robustness and flexible behavior invocation—already emerge naturally in BFMs trained with forward-backward (FB) representations and unsupervised reinforcement learning (URL)~\cite{bfm0}. FB learning enables a single learned representation to specify and retrieve
diverse behaviors through simple reward functions, making it well suited for
a range of interactive objectives. Meanwhile, URL optimizes for the FB objective rather than motion imitation, allowing the policy to prioritize task completion and recovery, and to deviate from reference motions when necessary. These properties motivate us to extend the same paradigm from general-purpose motion to physical interaction.

Based on this insight, we propose \method{}, which learns a unified, Interactive Behavioral Foundation Model through unsupervised RL with an FB representation.
It jointly embeds the humanoid's basic skills and interactive skills into a structured latent interaction space.
In interactive scenarios, the learned representation captures the coupled dynamics of the humanoid and objects, together with changes in their contact relationships.
By specifying simple high-level rewards, \method{} can invoke and compose diverse interactive behaviors, allowing the humanoid to perform skills such as carrying, pushing, and kicking objects, while quickly recovering from disturbances and interaction failures.
Specifically, our Local--Goal Objective Geometry Operator (LOGO) uses metric-conditioned local and goal intents with residual actor regularization to improve controllability and robustness.

\method{} consistently improves task success, object-placement accuracy, and disturbance recovery, while supporting skill chaining and controllable interaction styles. Real-world experiments further demonstrate diverse, robust interactions and rapid failure recovery (Fig. \ref{fig:teaser}). These results suggest that humanoid interaction can be learned as a reusable, object-centric, reward-addressable latent space rather than task-specific controllers or fixed motion plans. Our contributions are:

\begin{enumerate}[leftmargin=*, labelsep=0.5em]
\item We introduce \method{}, the first framework for building a promptable interactive BFM for general-purpose humanoid interaction. Using FB representations and URL, \method{} enables zero-shot execution of diverse interactive behaviors through a simple reward interface.

\item We incorporate object-contact information into the interaction state and LOGO as a local--goal metric-conditioned residual actor objective, improving controllability and robustness for humanoid-object interaction.

\item We demonstrate robust real-world humanoid behaviors with \method{}, including carrying, pushing, kicking, long-horizon behavior chaining, and recovery from large disturbances, interaction failures, and falls.
\end{enumerate}

\section{Related Works}

\subsection{Behavioral Foundation Models for Whole-body Control.}
Learning-based whole-body controllers have shown that motion priors can provide reusable control skills for humanoids, supporting diverse motion tracking and command-conditioned behaviors. Recent systems have further transferred these capabilities to real humanoids for whole-body tracking, teleoperation, and data collection~\cite{li2025hold,liao2025beyondmimic,li2025bfm}. SONIC~\cite{luo2025sonic} further illustrates the potential of scaling motion tracking and synthesis to broad humanoid behaviors. Building on this foundation, behavioral foundation models (BFMs)~\cite{metamotivo, bfm0, zeng2025behaviorfoundationmodelhumanoid}, seek a more scalable objective: learning a single general-purpose policy from large and diverse motion corpora that can be flexibly steered by latent behaviors, goals, or rewards. However, current BFMs remain largely body-centric, modeling humanoid motion without explicitly capturing objects and contact dynamics. In contrast, \method{} incorporates object and contact states, extending whole-body priors to persistent physical interaction.

\subsection{Humanoid--Object Interaction.}
Existing humanoid--object interaction methods commonly rely on teleoperation~\cite{luo2025sonic,zhang2025falcon,ze2025twist}, behavior cloning~\cite{he2025viral}, task-specific reward design~\cite{xue2025opening,su2025hitter,latent}, or dense motion-reference tracking~\cite{yang2025omniretarget,wang2026humanx,weng2025hdmi}. For example, HDMI~\cite{weng2025hdmi} learns interaction skills by co-tracking humanoid and object trajectories extracted from  videos. Although these methods produce high-quality motions, their dependence on reference trajectories limits adaptation after contact loss or unexpected object displacement. Reference-free methods such as PhysHSI~\cite{wang2025physhsi} and LessMimic~\cite{lin2026lessmimic} reduce reliance on motion tracking through task rewards or geometric interaction representations. OmniContact~\cite{omnicontact} improves flexibility by generating intermediate references online, but recovery requires additional planning and introduces latency. Rather than specializing inference for a particular interaction mode or replanning after failure, \method{} uses one object-aware behavioral policy to compose diverse behaviors, while directly reacting to contact disruptions within closed-loop control.

\subsection{Unsupervised RL and FB Representations.}
Unsupervised reinforcement learning aims to acquire reusable behaviors without task-specific policy training. Forward-backward representations provide a particularly flexible framework by embedding states, rewards, and policies into a shared latent space, allowing new behaviors to be invoked through simple reward specifications~\cite{onereptopall}. Prior FB-based models have demonstrated promptable locomotion and whole-body control, but have not explicitly modeled coupled humanoid--object dynamics~\cite{metamotivo,bfm0}. \method{} extends FB representations from body-only behavior generation to contact-rich physical interaction: its latent representation jointly incorporates humanoid and object states, enabling reward-conditioned task composition and recovery without retraining a task-specific policy or invoking online replanning.
\section{Preliminaries}

\subsection{Problem Formulation}
\begin{figure*}
    \centering
    \includegraphics[width=\linewidth]{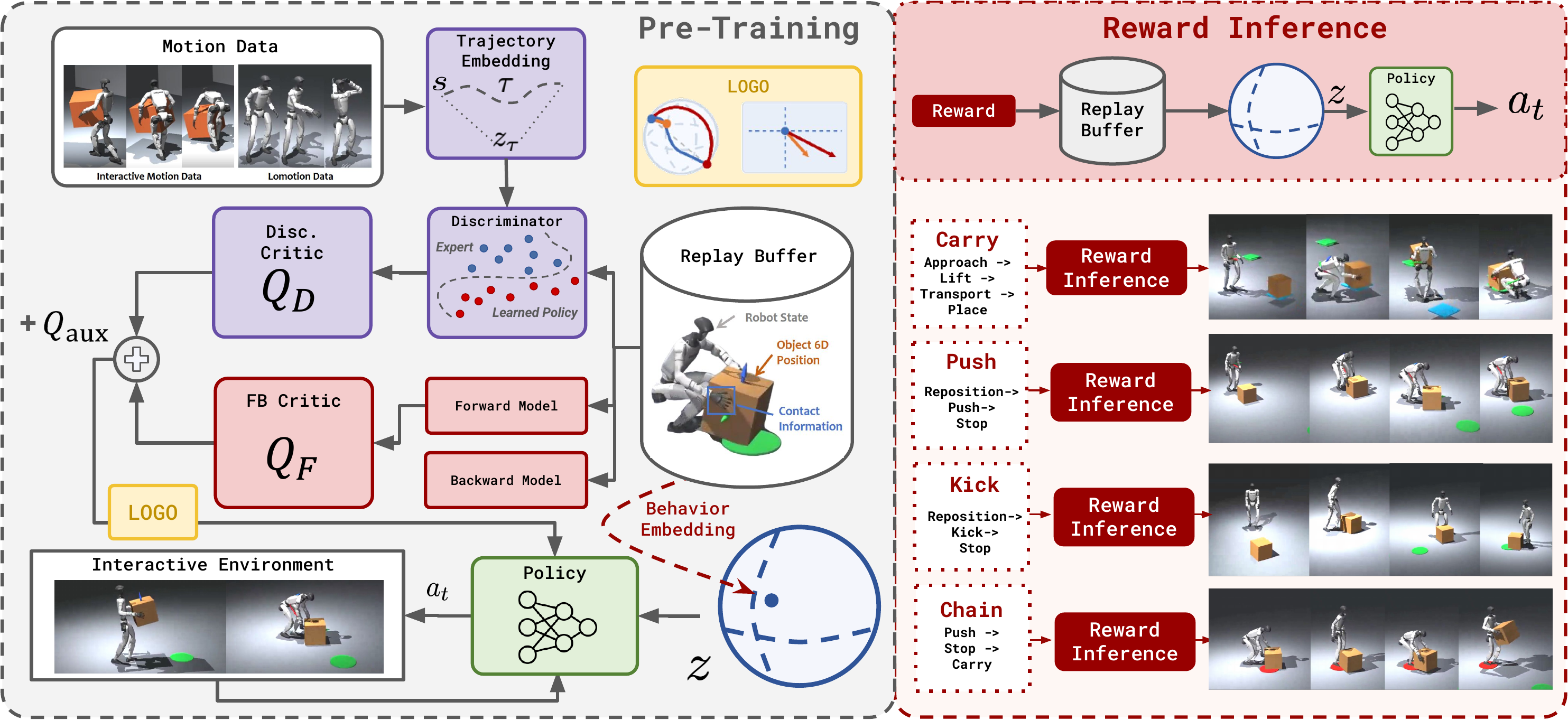}
    \caption{Overview of \method{}.
During pre-training, \method{} learns an interaction-aware behavioral embedding and a shared policy in an interactive environment, jointly optimized with forward-backward and discriminator critics. At inference, task rewards retrieve a corresponding latent from the learned embedding space, enabling the same policy to perform robust carrying, pushing, kicking, and compositional loco-manipulation behaviors.}
    \label{fig:overview}
        \vspace{-5mm}
\end{figure*}
We formulate humanoid-object interaction as a partially observable Markov
decision process $\mathcal{M}=(\mathcal{S},\mathcal{O},\mathcal{A},
P,\gamma)$, where $\mathcal{S}$, $\mathcal{O}$, and $\mathcal{A}$ denote the
state, observation, and action spaces, respectively, $P$ is the transition
dynamics, and $\gamma\in[0,1)$ is the discount factor. At time step $t$, the full interaction state is $s_t=(s_t^{\mathrm{hum}},s_t^{\mathrm{obj}}, s_t^{\mathrm{con}},s_t^{\mathrm{id}})$, where the four components describe the humanoid state, object states, contact information, and task identity, respectively. In this work, the carry-box, kick-box, and push-box behaviors are assigned the same task ID. The action
$a_t\in\mathcal{A}$ specifies joint-level targets executed by a low-level PD
controller. Since the full state is generally unavailable at deployment, the policy
conditions on a finite observation history
$h_t=(o_{t-H:t},a_{t-H:t-1})$, containing proprioceptive observations and
available object and contact measurements. During training, privileged state
information may be provided to the critic, while the actor only receives
deployable observations.

For a downstream reward $r:\mathcal{S}\rightarrow\mathbb{R}$, the objective is
$J_r(\pi)=\mathbb{E}_{\pi,P}[\sum_{t=0}^{\infty}\gamma^t r(s_t)]$. Our goal is
to learn a single latent-conditioned policy $\pi(a_t\mid h_t,z)$ that supports
diverse locomotion and interaction behaviors. Given a new task, its latent
command $z_r$ is inferred directly from the reward specification, enabling
behaviors such as carrying, pushing, and kicking without task-specific policy
optimization.

\subsection{Preliminaries on Forward-Backward Representations}

Forward-backward (FB) representations provide a reward-free framework for
learning a continuous family of policies. Let $\rho$ be a reference state
distribution. FB learning introduces a forward representation
$F:\mathcal{S}\times\mathcal{A}\times\mathbb{R}^{d}\rightarrow\mathbb{R}^{d}$
and a backward representation $B:\mathcal{S}\rightarrow\mathbb{R}^{d}$. Each
latent vector $z\in\mathbb{R}^{d}$ indexes a policy $\pi_z$. The discounted successor-state measure of $\pi_z$ is
$M^{\pi_z}(ds'\mid s,a)=\sum_{k=0}^{\infty}\gamma^k
\Pr(s_{t+k}\in ds'\mid s_t=s,a_t=a,\pi_z)$. FB learning approximates this
measure through the low-rank factorization
\begin{equation}
    M^{\pi_z}(ds'\mid s,a)
    \approx F(s,a,z)^\top B(s')\,\rho(ds').
    \label{eq:fb_factorization}
\end{equation}

The backward representation induces state features
$\phi(s)=(\mathbb{E}_{s'\sim\rho}[B(s')B(s')^\top])^{-1}B(s)$, such that a
latent $z$ defines the linear reward $r_z(s)=\phi(s)^\top z$. Under the FB
factorization, $F(s,a,z)^\top z$ approximates the action-value function of
$\pi_z$ for $r_z$. The corresponding policy is therefore given by
$\pi_z(s)\in\arg\max_{a\in\mathcal{A}}F(s,a,z)^\top z$.

For an arbitrary downstream reward $r$, its task representation can be
estimated without additional policy training:
\begin{equation}
    z_r
    =
    \mathbb{E}_{s\sim\rho}[B(s)r(s)]
    \approx
    \frac{1}{N}\sum_{i=1}^{N}B(s_i)r(s_i),
    \qquad s_i\sim\rho.
    \label{eq:reward_embedding}
\end{equation}
The resulting action-value function satisfies
$Q_r^{\pi_{z_r}}(s,a)\approx F(s,a,z_r)^\top z_r$. Goal reaching is a special
case for which the latent command can be obtained as $z_g=B(s_g)$.

Unlike conventional BFMs that model only humanoid motion, \method{} incorporates
object dynamics and contact states. Its shared FB representation therefore
supports rewards over object poses, contacts  through the
same interface.

\section{\method{}: Interactive Behavioral Foundation Model via Unsupervised RL}
\label{sec:method}
\method{} extends behavioral foundation models from body-centric motion control
to closed-loop humanoid--object interaction. As shown in
Fig.~\ref{fig:overview}, it learns a single latent-conditioned policy capable
of diverse locomotion and robust loco-manipulation behaviors. During
pre-training, \method{} combines online reward-free exploration in an
interactive environment with two offline datasets of unlabeled interaction and
locomotion motions to learn a compact humanoid--object behavioral space.
Contact information and complementary training objectives
promote robust interaction and stable contact, while LOGO provides local and
long-horizon metric conditioning. The resulting shared policy captures both
humanoid and object dynamics, enabling downstream rewards to invoke carrying,
pushing, kicking, and task composition without task-specific policy
optimization. Sec.~\ref{sec:pretraining} describes the pre-training procedure,
and Sec.~\ref{sec:inference} describes zero-shot reward inference.

\subsection{Pre-training of \method{}}
\label{sec:pretraining}

\paragraph{Interactive State and Contact Information}
To enable robust physical interaction and stable contact, \method{} extends
conventional body-centric BFMs to model the coupled interaction state
\begin{equation}
s =
\left(
s^{\mathrm{hum}},
s^{\mathrm{obj}},
s^{\mathrm{con}}
\right),
\end{equation}
which jointly captures humanoid motion, object dynamics, and contact
relationships.

The interactive state consists of $s_t^{\mathrm{hum}}$, $s_t^{\mathrm{obj}}$, and $s_t^{\mathrm{con}}$.
$s_t^{\mathrm{hum}}$ represents humanoid proprioception,
$s_t^{\mathrm{obj}}$ contains the object pose, velocity, goal displacement, and presence indicator,
and $s_t^{\mathrm{con}}$ contains bilateral hand--object contact indicators, hand positions, and hand-to-object displacement vectors.
The forward and backward representations are learned over this coupled interaction state. The actor receives deployable observation history
$h_t$ and a latent command $z$:
\begin{equation}
a_t \sim \pi_\vartheta(a_t \mid h_t,z).
\label{eq:interactive_policy}
\end{equation}
Consequently, states with similar humanoid configurations but different object
or contact information can induce different actions, enabling closed-loop interaction
and recovery after contact changes.

\noindent\paragraph{Pre-training Objective}

The pre-training BFM objective consists of three complementary components:

\noindent\textbf{Interaction-aware FB Representation.}
The FB component learns a reward-free representation of discounted future
interaction states. Its objective trains the forward representation
$F_\phi(s,a,z)$ and backward representation $B_\psi(s)$ to approximate the
successor-state measure in Eq.~\eqref{eq:fb_factorization}. Because the state
contains humanoid, object, and contact information, this objective organizes
the latent space according to both body and its physical effects on
objects. The resulting FB Q-function is
\begin{equation}
Q_F(s,a,z)=F_\phi(s,a,z)^\top z.
\label{eq:fb_q_function}
\end{equation}

\noindent\textbf{Conditional Discriminator.}
The discriminator learns to distinguish motion data from policy rollouts. Its
objective provides a motion-prior signal that encourages the policy to retain
human-like motions while learning interaction behaviors. The corresponding
style Q-function is denoted by $Q_D$.

\noindent\textbf{Auxiliary Interaction Critic.}
The auxiliary component learns an additional Q-function,
$Q_{\mathrm{aux}}$, from environment-level auxiliary rewards. It provides
stabilizing signals that promote smooth control and facilitate sim-to-real
transfer.
The three components are jointly optimized as
\begin{equation}
\mathcal{L}_{\mathrm{components}}
=
\mathcal{L}_{\mathrm{FB}}
+
\lambda_D \mathcal{L}_D
+
\lambda_{\mathrm{aux}}\mathcal{L}_{\mathrm{aux}}.
\label{eq:ibfm_component_objective}
\end{equation}
The shared policy is trained jointly with these components and is guided by
their complementary interaction, motion-prior, and auxiliary control signals.

\begin{figure}[!t]
    \centering
    \includegraphics[width=\columnwidth]{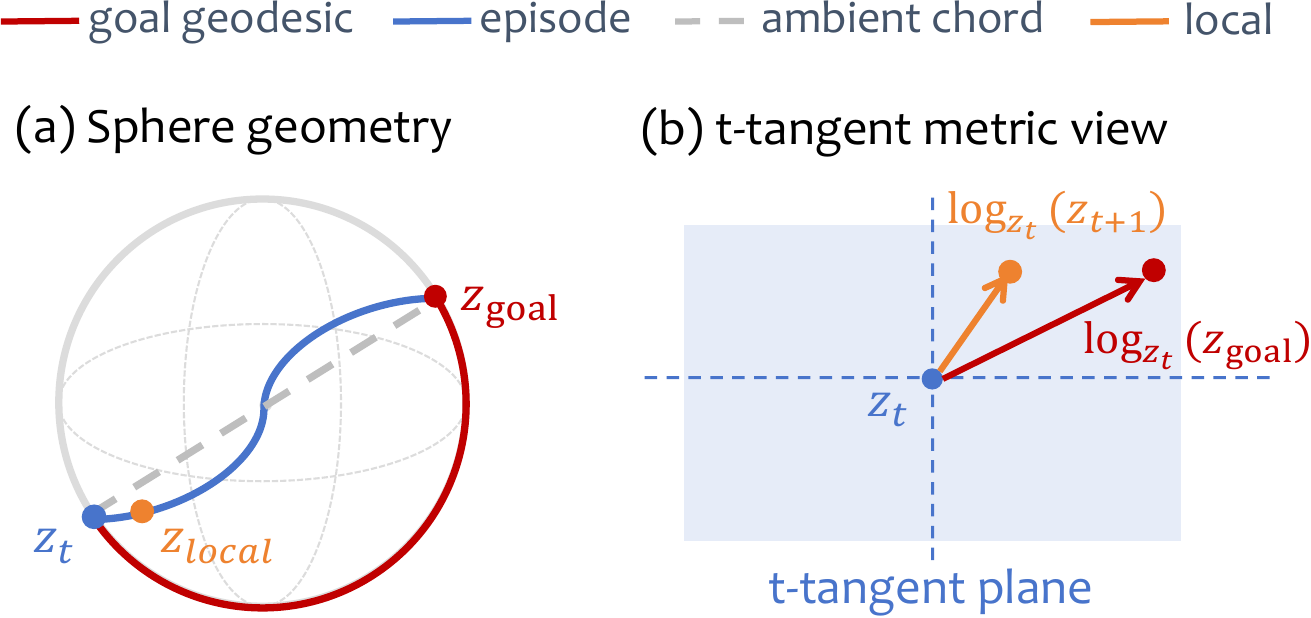}
    \caption{{LOGO metric conditioning.} The current latent is $z_t$;
    $z_{\mathrm{goal}}$ defines the long-horizon geodesic and
    $z_{\mathrm{local}}$ the immediate successor target. Their log-map vectors
    are evaluated in the same tangent plane and passed to the shared actor.}
    \label{fig:logo_logmap}
        \vspace{-5mm}
        \end{figure}

\paragraph{Joint Training on Interaction and Locomotion Data}
We jointly train the policy on locomotion and box-interaction data. Since the
locomotion data contain no box motion, all box-related entries in
$s_t^{\mathrm{obj}}$ and $s_t^{\mathrm{con}}$ are set to zero. The object-presence
indicator in $s_t^{\mathrm{obj}}$, together with $s_t^{\mathrm{id}}$, indicates
whether object interaction is required. For all box-interaction data, including
carrying, kicking, and pushing, we use the same task ID, allowing the policy
to learn a shared interaction representation instead of behavior-specific task
labels.
\paragraph{Local--Goal Objective Geometry Operator (LOGO)}
\label{sec:logo}
Physical interaction requires coordinating immediate contact-level responses
with longer-horizon objectives. LOGO captures these two scales by conditioning
a shared policy on local and goal-directed intents
(Fig.~\ref{fig:logo_logmap}). INTACT~\cite{sun2026intact} motivates this
local--goal interface by using one action model for physical transition and
future-goal intents; LOGO applies the idea to FB-based humanoid control with
spherical metric conditioning.

I-BFM without LOGO encodes a fixed eight-step window by averaging future backward
features.  Let $H=8$, let $B_\psi$ be the backward encoder, and let the latent
space be the radius-$r$ sphere $\mathbb{S}^{d-1}_r$.  We use the explicit
projection
\begin{align}
\mathcal{P}(u) &= r\frac{u}{\max(\lVert u\rVert_2,\epsilon)},
\qquad r=\sqrt{d}, \notag\\
z_t &= \mathcal{P}\!\left(B_\psi(s_t)\right).
\label{eq:logo_projection}
\end{align}
  The original command and the two metric
targets are
\begin{align}
z_{\mathrm{no\text{-}LOGO}} &=
\mathcal{P}\!\left(\frac{1}{H}\sum_{k=1}^{H}B_\psi(s_{t+k})\right), \notag\\
z_{\mathrm{local}} &= \mathcal{P}\!\left(B_\psi(s_{t+1})\right), \notag\\
z_{\mathrm{goal}} &= \mathcal{P}\!\left(B_\psi(s_{t+H})\right).
\label{eq:logo_targets}
\end{align}

The average can blur states with different first actions; the local successor
retains immediate physical intent, while the endpoint retains task intent.

For a target $z_\star\in\{z_{\mathrm{local}},z_{\mathrm{goal}}\}$, the spherical
log-map gives a tangent-space metric vector.  With
$x=z_t/r$, $y=z_\star/r$, $v=y-(x^\top y)x$, and
$\theta=\operatorname{atan2}(\lVert v\rVert_2,x^\top y)$,
\begin{equation}
    m_t(z_\star)=\operatorname{Log}_{z_t}(z_\star)
    =r\theta\frac{v}{\max(\lVert v\rVert_2,\epsilon)}.
\label{eq:logo_logmap}
\end{equation}
Thus the magnitude records geodesic distance and the direction records the
locally correct direction on the latent sphere; local and goal vectors need not
be collinear.

LOGO uses one shared actor for local and goal intents:
\begin{equation}
a_t(z_\star)\sim\pi_\vartheta\!\left(a\mid h_t,m_t(z_\star)\right).
\label{eq:logo_actor}
\end{equation}
Here $h_t$ is the deployable observation history. The log-map replaces the actor's latent command; the FB and style critics retain
canonical spherical commands. For actions sampled as in Eq.~\eqref{eq:logo_actor}, the native actor loss is
\begin{align}
Q_{\mathrm{native}}(s,a,z) &= Q_F(s,a,z) \notag\\
&\enspace +w\!\left(\lambda_D Q_D(s,a,z)
+\lambda_{\mathrm{aux}}Q_{\mathrm{aux}}(s,a,z)\right), \notag\\
\ell_{\mathrm{native}}(s,z) &=
-\mathbb{E}_{a}
\left[Q_{\mathrm{native}}(s,a,z)\right].
\label{eq:logo_native_actor}
\end{align}
The critics use the original pessimistic aggregation; $w$ is the detached
batch mean of $|Q_F|$ when native regularization scaling is enabled.
For an episode-aligned replay clip $C=(s_t,\ldots,s_{t+H})$, both targets
are evaluated at the same anchor:
\begin{align}
\mathcal{L}_{\mathrm{local}} &=
\mathbb{E}_{C}\left[\ell_{\mathrm{native}}(s_t,z_{\mathrm{local}})\right], \notag\\
\mathcal{L}_{\mathrm{goal}} &=
\mathbb{E}_{C}\left[\ell_{\mathrm{native}}(s_t,z_{\mathrm{goal}})\right], \notag\\
\mathcal{L}_{\mathrm{LOGO}} &= \mathcal{L}_{\mathrm{base}}+
\beta\!\left(\frac{4}{7}\mathcal{L}_{\mathrm{local}}+
\frac{3}{7}\mathcal{L}_{\mathrm{goal}}\right).
\label{eq:logo_objective}
\end{align}
$\mathcal{L}_{\mathrm{base}}$ retains the original mixed-command distribution
and native actor objective. The $4{:}3$ ratio and $\beta=0.005$ were selected
by a controlled hyperparameter sweep. Backward embeddings are stop-gradient
for these actor updates. LOGO adds a local--goal loss residual, not an action-label
loss or a sum of latent vectors; the discriminator, FB representation losses,
and reward-based latent inference remain unchanged.

\subsection{Inference of \method{}}
\label{sec:inference}

\paragraph{Reward Inference}
For a downstream reward $r$, we infer its latent command using the learned
backward representation:
\begin{equation}
z_r
=
\mathcal{P}\left(
\frac{1}{N}\sum_{i=1}^{N}B_\psi(s_i)r(s_i)
\right),
\qquad s_i\sim\rho.
\label{eq:ibfm_reward_inference}
\end{equation}
The shared policy then executes $a_t\sim\pi_\vartheta(a_t\mid h_t,z_r)$ in
closed loop, without task-specific policy optimization.

\paragraph{Task Rewards}
We decompose each long-horizon interaction into phases and construct a reward
for each phase by combining contact mode, interaction geometry, object motion,
and stability. Shared locomotion rewards produce approach and turning latents.
For Carry Box, phase rewards encourage grasping and lifting, goal-directed 3D
transport, controlled lowering, release, and stable placement. For Push Box,
they encourage rear hand contact, grounded planar transport,
and near-goal stopping. For Kick Box, they select aligned, hand-free impacts
that increase the box velocity toward the goal while preserving balance and
box stability. Reward-weighted inference over replay transitions produces a
latent command for each phase, which is selected according to object distance,
contact, and task progress.
\section{Experiments}

\begin{table*}[t]
\centering
\caption{
Quantitative comparison across three loco-manipulation tasks.
SR is the success rate, $E_{\mathrm{obj}}$ is the final
object-to-goal error in meters (clipped at
$1$ before aggregation), and SR-O and SR-R denote success
rates under object-fall and robot-fall perturbations, respectively.
A dash ($-$) denotes that the corresponding method does not support the task.}
\label{tab:main_results}
\setlength{\tabcolsep}{4.5pt}
\renewcommand{\arraystretch}{1.15}
\resizebox{\textwidth}{!}{
\begin{tabular}{lcccccccccccc}
\toprule
\multirow{2}{*}{\textbf{Method}}
& \multicolumn{4}{c}{\textbf{Carry Box}}
& \multicolumn{4}{c}{\textbf{Push Box}}
& \multicolumn{4}{c}{\textbf{Kick Box}} \\
\cmidrule(lr){2-5}
\cmidrule(lr){6-9}
\cmidrule(lr){10-13}
& \textbf{SR $\uparrow$}
& \textbf{$E_{\mathrm{obj}}$ $\downarrow$}
& \textbf{SR-O $\uparrow$}
& \textbf{SR-R $\uparrow$}
& \textbf{SR $\uparrow$}
& \textbf{$E_{\mathrm{obj}}$ $\downarrow$}
& \textbf{SR-O $\uparrow$}
& \textbf{SR-R $\uparrow$}
& \textbf{SR $\uparrow$}
& \textbf{$E_{\mathrm{obj}}$ $\downarrow$}
& \textbf{SR-O $\uparrow$}
& \textbf{SR-R $\uparrow$} \\
\midrule

PhysHSI~\cite{wang2025physhsi}
& $37.7$\std{2.3}\% & $0.63$\std{0.03}
& $1.0$\std{0.0}\% & $0.0$\std{0.0}\%
& -- & -- & -- & --
& -- & -- & -- & -- \\

LessMimic~\cite{lin2026lessmimic}
& $26.0$\std{1.7}\% & $0.79$\std{0.01}
& $7.0$\std{2.6}\% & $2.7$\std{2.1}\%
& $1.7$\std{1.5}\% & $0.95$\std{0.01}
& $1.0$\std{1.0}\% & $0.7$\std{0.6}\%
& -- & -- & -- & -- \\

OmniContact~\cite{omnicontact}
& $92.7$\std{0.6}\% & $0.19$\std{0.01}
& $60.0$\std{7.0}\% & $1.3$\std{2.3}\%
& $42.0$\std{2.6}\% & \best{$0.24$\std{0.01}}
& $40.0$\std{10.5}\% & $2.7$\std{1.5}\%
& $1.3$\std{0.6}\% & $0.96$\std{0.02}
& $0.0$\std{0.0}\% & $0.0$\std{0.0}\% \\

\midrule
\rowcolor{gray!10}
\textbf{\method{} (Ours)}
& \best{$94.3$\std{3.1}\%}
& \best{$0.15$\std{0.02}}
& \best{$91.0$\std{1.0}\%}
& \best{$89.3$\std{1.5}\%}
& \best{$75.0$\std{6.6}\%}
& $0.31$\std{0.05}
& \best{$46.0$\std{4.4}\%}
& \best{$74.3$\std{5.8}\%}
& \best{$70.0$\std{3.6}\%}
& \best{$0.28$\std{0.03}}
& \best{$57.3$\std{7.6}\%}
& \best{$60.0$\std{8.7}\%} \\
\bottomrule
\end{tabular}
}
\end{table*}
\begin{figure*}[ht]
    \centering
    \includegraphics[width=\linewidth]{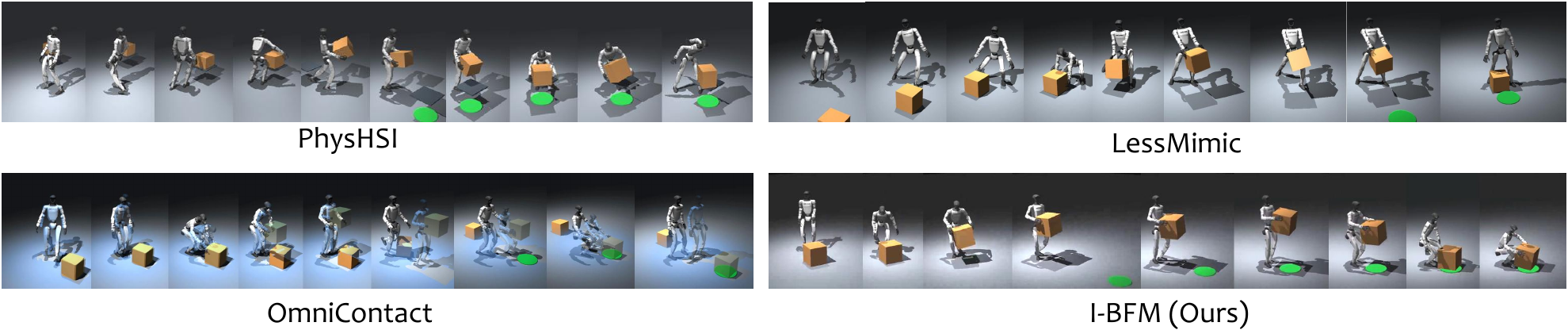}
  \caption{{Qualitative Results.}
\method{} reliably completes box transport, while baselines are less consistent.}
    \label{fig:carry_compare_to_baseline}
    \vspace{-2mm}
\end{figure*}

\begin{figure*}[h]
    \centering
    \includegraphics[width=\linewidth]{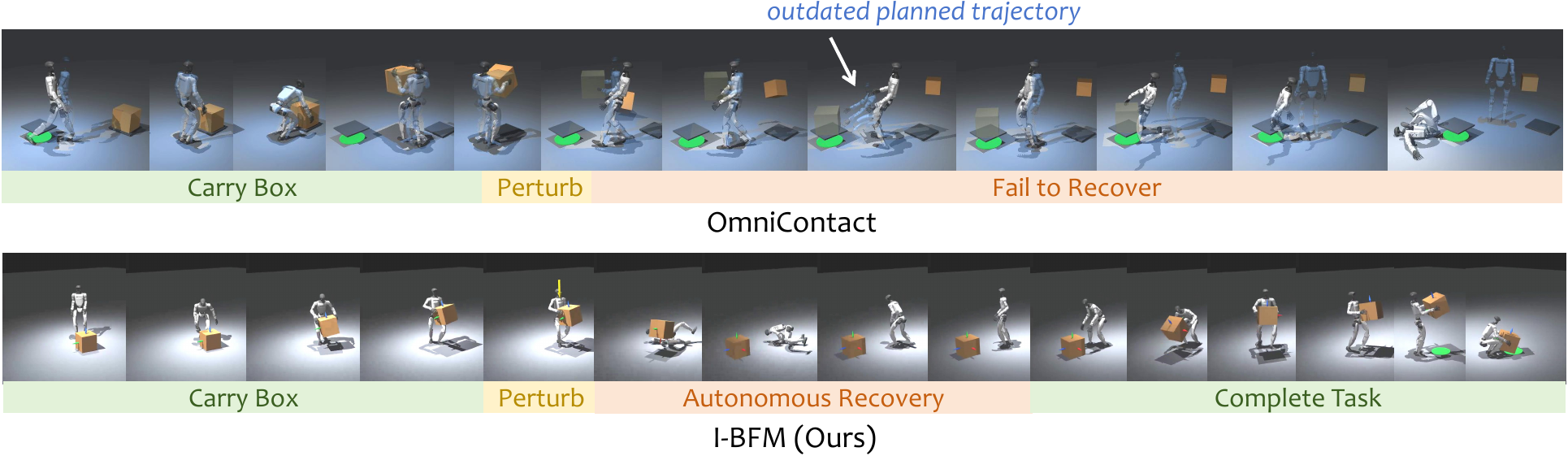}
    \caption{\textbf{Recovery from Perturbations.}
    During object perturbation, OmniContact fails to follow the replanned
    trajectory (translucent blue), while \method{} autonomously recovers and
    completes the task even under robot perturbation.}
    \label{fig:recover}
    \vspace{-5mm}
\end{figure*}

To comprehensively evaluate \method{}, we design experiments to answer the following questions:
(i) how well \method{} performs on interactive loco-manipulation tasks \emph{compared with prior baselines} (Sec.\ref{sec:main_results});
(ii) whether \method{} is robust to external disturbances (Sec.\ref{sec:robustness});
(iii) how much \emph{LOGO design} contributes to its performance (Sec.\ref{sec:ablation});
(iv) whether the learned interface generalizes to \emph{other control modes}, such as motion tracking and goal reaching (Sec.\ref{sec:other_interface});
and (v) how well \method{} performs in the \emph{real world}, particularly in terms of robustness (Sec.\ref{sec:real}). To further demonstrate the potential of \method{}, we evaluate its ability among multiple downstream loco-manipulation behaviors in Sec.\ref{sec:downstream}.

\subsection{Experimental Setup}

\textbf{Tasks.} We evaluate \method{} on three humanoid loco-manipulation tasks:
\textit{Carry Box}, which requires lifting and transporting a box;
\textit{Push Box}, which requires pushing a box to a goal; and
\textit{Kick Box}, which requires kicking a box toward a target. Together,
they cover hand-based transport, sustained pushing, and foot--object interaction.

\textbf{Randomization and Metrics.}
We evaluate each method with randomized robot and object initializations and task targets.
We report success rate and final object-to-goal error, and additionally measure robustness under object-fall and robot-fall perturbations, denoted SR-O and SR-R, respectively.

\textbf{Baselines.}
We compare \method{} with PhysHSI~\cite{wang2025physhsi},
LessMimic~\cite{lin2026lessmimic}, and OmniContact~\cite{omnicontact} in
MuJoCo using the G1 humanoid. PhysHSI uses adversarial motion-prior RL for
humanoid-scene interaction. LessMimic combines distance-field geometric
representations with latent interaction modeling. OmniContact uses
contact-flow-conditioned meta-skills for long-horizon loco-manipulation.

\subsection{Main Task Performance}
\label{sec:main_results}

To evaluate interactive loco-manipulation, we compare \method{} with prior humanoid interaction methods on three tasks in Table~\ref{tab:main_results}. On Carry Box, \method{} achieves a $94.3\%$ success rate and the lowest final object error of $0.11$ m. This accuracy highlights its object-centric representation, which directly optimizes object-level task rewards. In contrast, OmniContact suffers from tracking errors and residual inaccuracies not corrected by re-planning. By jointly learning interaction representations and low-level control, \method{} avoids this separation and achieves higher precision.

\method{} also performs well on the other two loco-manipulation tasks, demonstrating that the learned representation is not limited to a single interaction pattern.
Because our method is trained with unsupervised RL, it explores diverse states and interaction outcomes, allowing new interactive tasks to be instantiated rapidly with minimal task-specific data or tuning.
Notably, our training data does not include demonstrations of carrying boxes from one height to a target with a different height; thus, the height-offset Carry Box setting evaluates zero-shot generalization beyond the demonstrated interaction distribution.
This further validates the generalization benefits brought by the combination of FB and unsupervised RL.
Qualitatively (Fig.~\ref{fig:carry_compare_to_baseline}), \method{} produces natural, human-like behaviors comparable to PhysHSI and OmniContact.
At the same time, it supports controllable interactive behavior patterns (Sec.\ref{sec:downstream}) and achieves substantially better recovery under object and robot perturbations (Sec.\ref{sec:robustness}).
We attribute these properties to the FB-based representation, which enables interactive behaviors to be specified by high-level rewards and executed through a unified low-level controller.

\subsection{Robustness and Recovery}
\label{sec:robustness}

We further evaluate whether \method{} is robust to external disturbances, including object displacement and robot falls. During rollout, we perturb either the object or the robot and measure whether the policy can recover and still complete the task.
As shown in Table~\ref{tab:main_results}, \method{} maintains strong recovery performance with little degradation from the nominal setting. In contrast, OmniContact drops to $60.0\%$ SR-O and $1.3\%$ SR-R, while other baselines show even weaker recovery.

Qualitatively, as shown in Fig.~\ref{fig:recover}, after an object-fall perturbation, the robot may enter a state from which it cannot feasibly transition to the state assumed by the replanned trajectory. Even when replanning eventually finds a feasible recovery trajectory, the planning and execution delay can prevent timely recovery. Robot-fall recovery is even more challenging for methods that rely on nominal motion plans. In contrast, through broad state-space exploration during unsupervised RL, \method{} continues optimizing the same high-level task objective despite large state deviations, without requiring alignment with a predefined trajectory. Recovery behaviors therefore emerge naturally within the unified policy, demonstrating the advantage of reward-specified interactive skills over fixed reference trajectories.

\subsection{Ablation Study}
\label{sec:ablation}

Table~\ref{tab:ablation} evaluates the contributions of LOGO in \method{}. In \emph{\method{} w/o LOGO}, we remove the local--goal metric conditioning and instead condition the actor on the averaged features over an eight-step window. All other training objectives and model components remain unchanged. We report the success rate (SR), final object-to-goal error ($E_{\mathrm{obj}}$), and joint-position and velocity tracking errors ($E_{\mathrm{jpos}}$ and $E_{\mathrm{vel}}$) to assess task performance and motion quality.

We find that LOGO substantially improves SR while only slightly reducing tracking errors. This suggests that its benefit does not primarily arise from improved motion imitation. Instead, LOGO resolves the ambiguity caused by averaging future features: local metric conditioning preserves the immediately feasible interaction intent, while goal conditioning maintains long-horizon task progress, leading to more consistent contact transitions and task completion.

\begin{table}[h]
\centering
\small
\caption{Ablation study of \method{}.}
\label{tab:ablation}
\resizebox{\linewidth}{!}{
\begin{tabular}{lcccc}
\toprule
\textbf{Method}
& \textbf{SR $\uparrow$}
& $\boldsymbol{E_{\mathrm{obj}}} \downarrow$
& $\boldsymbol{E_{\mathrm{jpos}}} \downarrow$
& $\boldsymbol{E_{\mathrm{vel}}} \downarrow$ \\
\midrule
\method{} w/o LOGO
& $51.0\%$
& $0.33$
& $1.06$
& $129.7$ \\

\rowcolor{gray!10}
\method{}
& \best{$94.3\%$}
& \best{$0.15$}
& \best{$1.02$}
& \best{$126.9$} \\
\bottomrule
\end{tabular}
}
\end{table}

\vspace{-3mm}
\subsection{Other Interfaces and Control Mode}
\label{sec:other_interface}
\vspace{-3mm}

\begin{figure}[h]
    \centering
    \includegraphics[width=\linewidth]{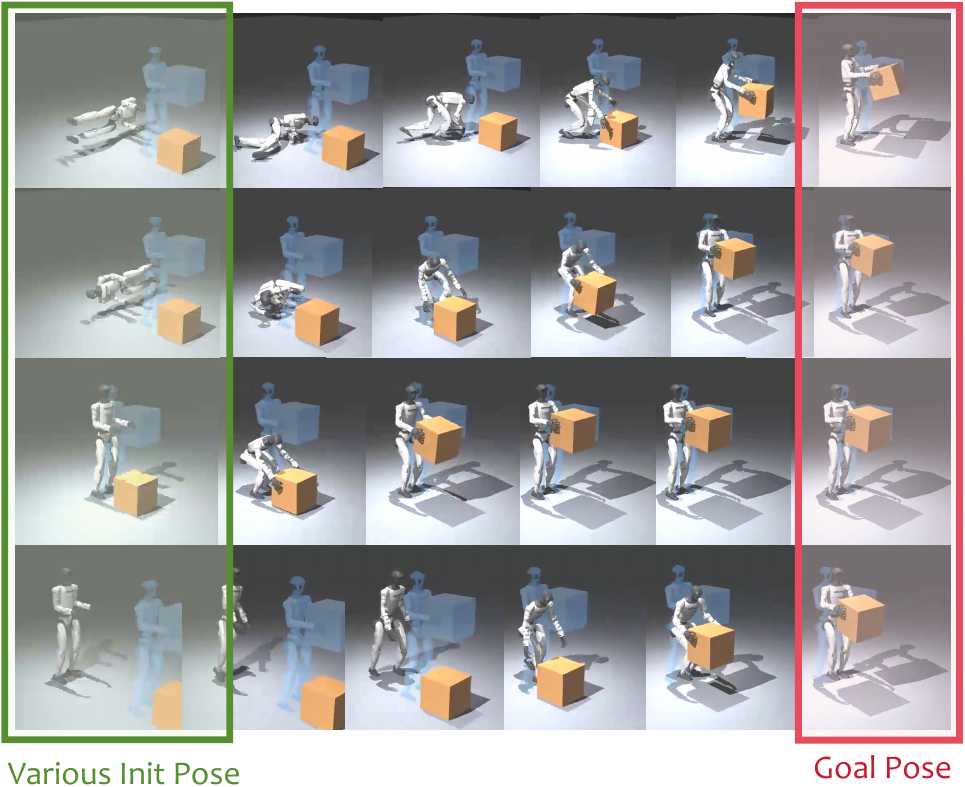}
    \caption{{Goal Reaching from diverse initial states.}
From diverse robot poses and box configurations (green), a single \method{} policy recovers and transports the box to the shared goal pose (red). Translucent blue figures show the goal pose.}
    \label{fig:carry_box_goal_reaching}
\end{figure}

\begin{figure}[h]
    \centering
    \includegraphics[width=\linewidth]{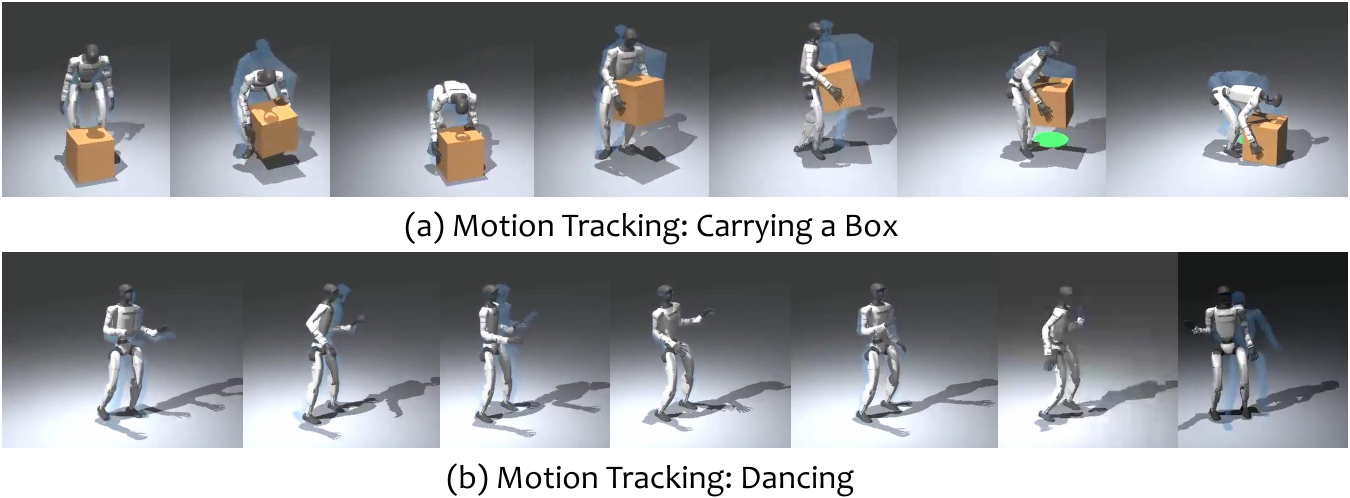}
    \caption{{Motion tracking with a unified policy.}
\method{} tracks generated whole-body motions for both box carrying (top) and dancing (bottom). Translucent blue figures denote the reference motion.}
    \label{fig:tracking}
    \vspace{-5mm}
\end{figure}

Beyond reward-conditioned meta-skills, we examine whether \method{} supports goal reaching and motion tracking without task-specific policy optimization. For Carry Box, under goal reaching, the target object state serves as the goal and the corresponding latent command is retrieved from the learned representation. As shown in Fig.~\ref{fig:carry_box_goal_reaching}, the robot transports the box between arbitrary locations by composing behaviors such as approaching, lifting, carrying, placing, and getting up. Under motion tracking, latent commands derived from reference trajectories enable the same policy to reproduce interaction motions while maintaining coordinated humanoid--object contact.

Beyond object interaction, \method{} also retains individual motor skills, accurately tracking diverse whole-body motions such as dancing (Fig.~\ref{fig:tracking}(b)). These results show that the unified interaction-aware representation supports reward inference, goal reaching, and motion tracking while preserving diverse individual and interactive behaviors.

\begin{figure*}[h]
    \centering
    \includegraphics[width=\linewidth]{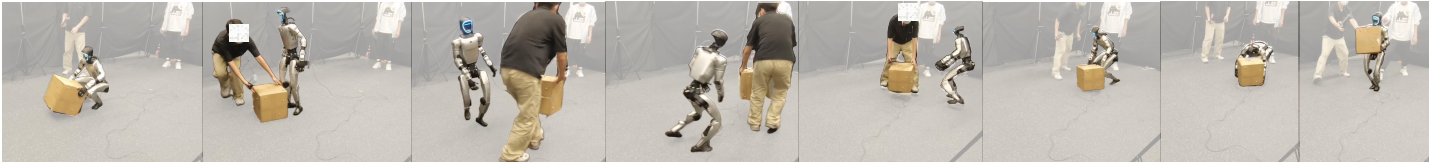}
\caption{{Timely Adaptation to Box Displacement.}
When the box is moved during execution, \method{} promptly updates its motion, re-approaches the new box pose, and resumes transportation in closed loop.}
    \label{fig:real}
        \vspace{-2mm}
\end{figure*}

\begin{figure*}[h]
    \centering
    \includegraphics[width=\linewidth]{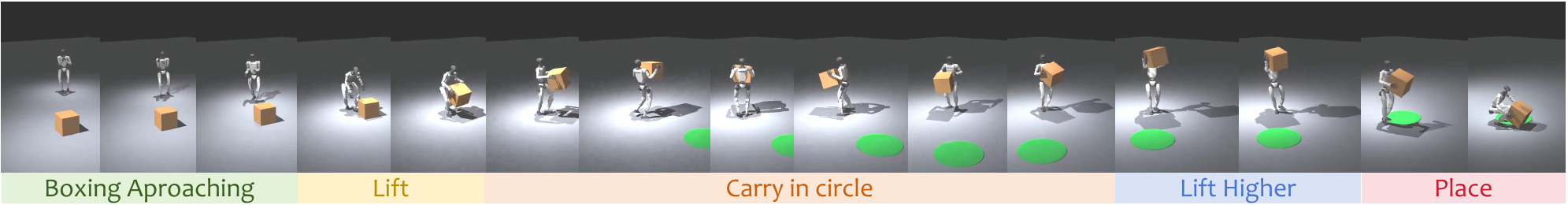}
    \caption{{Compositional box manipulation.}
A single \method{} policy sequentially approaches, lifts, carries, raises, and places the box in response to changing task commands.}
    \label{fig:Stylistic}
    \vspace{-5mm}
\end{figure*}

\subsection{Real-world Validation}
\label{sec:real}
We deploy \method{} on a Unitree G1 humanoid robot in a motion-capture workspace. We use a $0.35\,\mathrm{m}\times0.35\,\mathrm{m}\times0.35\,\mathrm{m}$ box
weighing $0.7\,\mathrm{kg}$, and execute the policy at $50\,\mathrm{Hz}$. Despite the sim-to-real gap, the same policy smoothly completes
diverse interactive tasks on hardware with only minor
degradation from simulation, without task-specific retraining or online
replanning. 

Fig.~\ref{fig:teaser} presents representative real-world rollouts. In
Fig.~\ref{fig:teaser}(a), the robot recovers from an unsuccessful box-handling
attempt and retries the task. In Fig.~\ref{fig:teaser}(b), the robot is
externally perturbed during carrying and the box is displaced; it regains
balance, re-approaches the box, and resumes transport.
Figs.~\ref{fig:teaser}(c) and \ref{fig:teaser}(d) demonstrate pushing and
kicking, respectively. Fig.~\ref{fig:real} shows timely adaptation when the box is displaced
during execution. Together, these results demonstrate diverse whole-body
interaction, goal-conditioned control, and closed-loop recovery on real hardware
using a single shared policy.

\subsection{Downstream Tasks}
\label{sec:downstream}

\textbf{Stylistic Interactive Behaviors.} 
Fig.~\ref{fig:Stylistic} shows how \method{} combines stylistic motion and box manipulation within a single policy. The controller blends a boxing latent with the approach latent, then uses a reward-inferred lift latent to grasp the box. During transport, a turn--boxing latent blend produces a full carried-box rotation, followed by a blend with an overhead-lifting latent. Contact, yaw, and heading feedback govern these transitions before the robot lowers and places the box. The same actor executes every phase without policy switching or full-body trajectory tracking.

\begin{figure}[h]
    \centering
    \includegraphics[width=\linewidth]{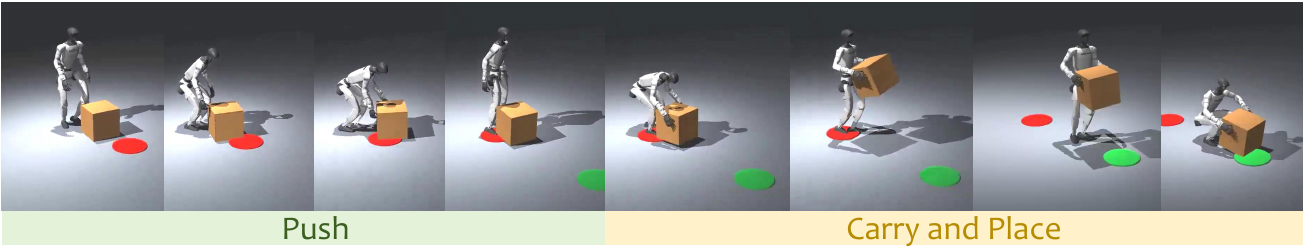}
    \caption{{Task Chaining.}
The robot sequentially pushes, carries, and places the box.}
    \label{fig:chain}
\end{figure}

\textbf{Skill Chaining} For multi-stage tasks, we sequentially update the reward specification when
each subgoal is completed. The resulting latent commands invoke the required
behaviors through the same shared policy, enabling long-horizon task
composition, such as push-carry-place, without policy switching or replanning (Fig.~\ref{fig:chain}).

\vspace{-5mm}
\section{Conclusion}
We introduced \method{}, an interactive behavioral foundation model that extends the generality and robustness of BFMs from humanoid motion to physical interaction. By learning an object-aware FB representation with unsupervised RL, \method{} provides a unified reward interface for diverse interaction behaviors without task-specific controllers or motion plans. Across simulation and real-world experiments, \method{} enables versatile humanoid-object interaction, flexible behavior invocation, and robust recovery from failures. We hope this work provides a step toward general-purpose humanoids that can interact with the physical world through a reusable and robust behavioral representation.

\bibliographystyle{IEEEtran}
\bibliography{library}

\useRomanappendicesfalse
\appendices
\section{Evaluation Protocols}
\label{details_eval}
\paragraph{Carry Box}
Each method is evaluated in MuJoCo environment. Box dimensions and mass are set to the midpoints of the corresponding training ranges specified in each method's codebase; parameters fixed during training retain their training values. The robot starts from a standing pose at $(0,0)$ with heading sampled uniformly from $[-\pi,\pi]$. The initial box center is sampled with $x,y\sim\mathcal{U}(-5,5)$ and $z\sim\mathcal{U}(0.15,0.8)$. The target center is $(1,1,z_g)$, where $z_g\sim\mathcal{U}(0.15,0.8)$. Two support platforms, each measuring $0.4\times0.4\times0.02$\,m, are placed beneath the initial and target box poses, with their top surfaces aligned with the corresponding box bottoms. Box--platform collisions are enabled, while robot--platform collisions are disabled.

\paragraph{Push Box}
The training environments differ across methods, including whether the box is wheeled. To standardize the evaluation object, we use a non-wheeled box with a mass of 0.7\,kg and side lengths of approximately 0.4\,m. In our tests, using OmniContact's original box did not yield a higher success rate than using this standardized box. Under the shared initialization protocol, the robot starts from a standing pose at $(0,0)$ with heading sampled uniformly from $[-\pi,\pi]$. The initial box XY position is sampled uniformly from $[-5,5]^2$, and the target XY position is fixed at $(1,1)$. The box rests on the ground, with both its initial and target center heights set to half its height. LessMimic Push is evaluated separately using its native box and initialization due to inference limitations: the robot starts at $(0,0)$ with fixed heading, the box center is initialized at $(0.4,0,h/2)$, and the target XY position is sampled uniformly from $[-5,5]^2$, where $h$ denotes box height. Its results are supplementary and not directly comparable under the shared protocol.

\paragraph{Replanning}
Replanning is enabled wherever supported by the native inference implementation. OmniContact uses its native replanning mechanism. Methods without a separate replanning module retain their native closed-loop controllers.

\paragraph{Metrics and Termination}
The maximum episode duration is 60\,s, corresponding to 12,000 simulation steps at a timestep of 0.005\,s. Success requires an object-center-to-target distance below 0.2\,m. Carry Box uses 3D distance, except for LessMimic, which uses XY distance. For the reported Push Box results, LessMimic uses XY distance and OmniContact uses 3D distance.

Carry Box and LessMimic Push episodes terminate early and are counted as successful when the distance criterion holds for 100 consecutive simulation steps; otherwise, success is determined at the time limit. The reported OmniContact Push evaluation runs for the full 60\,s and determines success at the final step. We report success rate (SR) and final 3D object-to-target error ($E_{\mathrm{obj}}^{T}$), measured at episode termination.

\paragraph{Repetitions}
Each evaluation comprises three runs of 100 episodes. The updated OmniContact Carry Box evaluation and the LessMimic and PhysHSI evaluations use seed blocks 0--99, 100--199, and 200--299. The reported OmniContact Push evaluation uses three blocks of 100 consecutive seeds starting at 166559744, 266559744, and 366559744. We report the mean and sample standard deviation of the three run-level success rates and mean final errors.

\paragraph{Perturbations}
We report success rates under object-fall and robot-fall perturbations, denoted by SR-O and SR-R, respectively. A perturbation is triggered once the box reaches the midpoint of its transport progress, i.e., when its XY distance to the target becomes smaller than half of its initial XY distance to the target. For Carry Box, SR-O applies a vertical downward force of 200\,N to the box for 0.2\,s, while SR-R applies a vertical downward force of 2,000\,N to the robot pelvis for 0.3\,s, causing the robot to fall. For Push Box and Kick Box, we instead apply the corresponding forces in a uniformly sampled horizontal direction and restrict the perturbed box position to remain within 2\,m of its position in the unperturbed rollout. The success criterion is unchanged from the nominal evaluation.

\end{document}